\pdfoutput=1

\documentclass[11pt]{article}

\usepackage{acl}

\usepackage{times}
\usepackage{latexsym}
\usepackage{amsmath}
\usepackage{amssymb}
\usepackage{verbatim}
\usepackage{booktabs}
\usepackage{comment}
\usepackage{pifont}
\usepackage{cancel}
\usepackage[normalem]{ulem}
\usepackage{makecell}
\usepackage{multirow}

\usepackage[T1]{fontenc}

\usepackage[utf8]{inputenc}

\usepackage{microtype}
\usepackage{graphicx}

\usepackage{inconsolata}

\title{Multi-matrix Factorization Attention}

\author{
Jingcheng Hu$^{1,2}$\thanks{Equal contribution.}\thanks{Work done during an internship at StepFun.},
Houyi Li$^{1,3}$\footnote[1]{},
Yinmin Zhang$^{1}$,
Zili Wang$^{1}$,
Shuigeng Zhou$^{3}$,
\\
\textbf{%
Xiangyu Zhang$^{1,4}$,
Heung-Yeung Shum$^{2}$,
Daxin Jiang$^{1}$
}\\\\
$^1$~StepFun \qquad
$^2$~Tsinghua University
 \\$^3$~Fudan University \qquad
 $^4$~Megvii Technology
}

\begin{document}
\maketitle

\begin{abstract}

We propose novel attention architectures, Multi-matrix Factorization Attention (MFA) and MFA-Key-Reuse (MFA-KR). 
Existing variants for standard Multi-Head Attention (MHA), including SOTA methods like MLA, fail to maintain as strong performance under stringent Key-Value cache (KV cache) constraints.
MFA enhances model capacity by efficiently scaling up both the number and dimension of attention heads through low-rank matrix factorization in the Query-Key (QK) circuit. 
Extending MFA, MFA-KR further reduces memory requirements by repurposing the key cache as value through value projection re-parameterization. 
MFA's design enables strong model capacity when working under tight KV cache budget, while MFA-KR is suitable for even harsher KV cache limits with minor performance trade-off. 
Notably, in our extensive and large-scale experiments, the proposed architecture outperforms MLA and performs comparably to MHA, while reducing KV cache usage by up to 56\% and 93.7\%, respectively.

\end{abstract}

\section{Introduction}

The decoder-only transformer with standard Multi-Head Attention (MHA)  \cite{vaswaniAttentionAllYou2017,radford2018improving} has become the de facto architecture for large language models. 
Its autoregressive nature enables the reuse of cached attention key-value tensors (KV cache) from previous tokens, significantly relieving the computation overhead during the step-by-step decoding \cite{pope2023efficiently}. 
However, the KV cache memory footprint scales linearly with both batch size and sequence length, leading to large amount of memory occupancy and traffic, which becomes the primary bottleneck during the decoding phase of LLM \cite{yuan2024llm}. 

\begin{figure}[t!]
    \centering
    \includegraphics[width=0.99\linewidth]{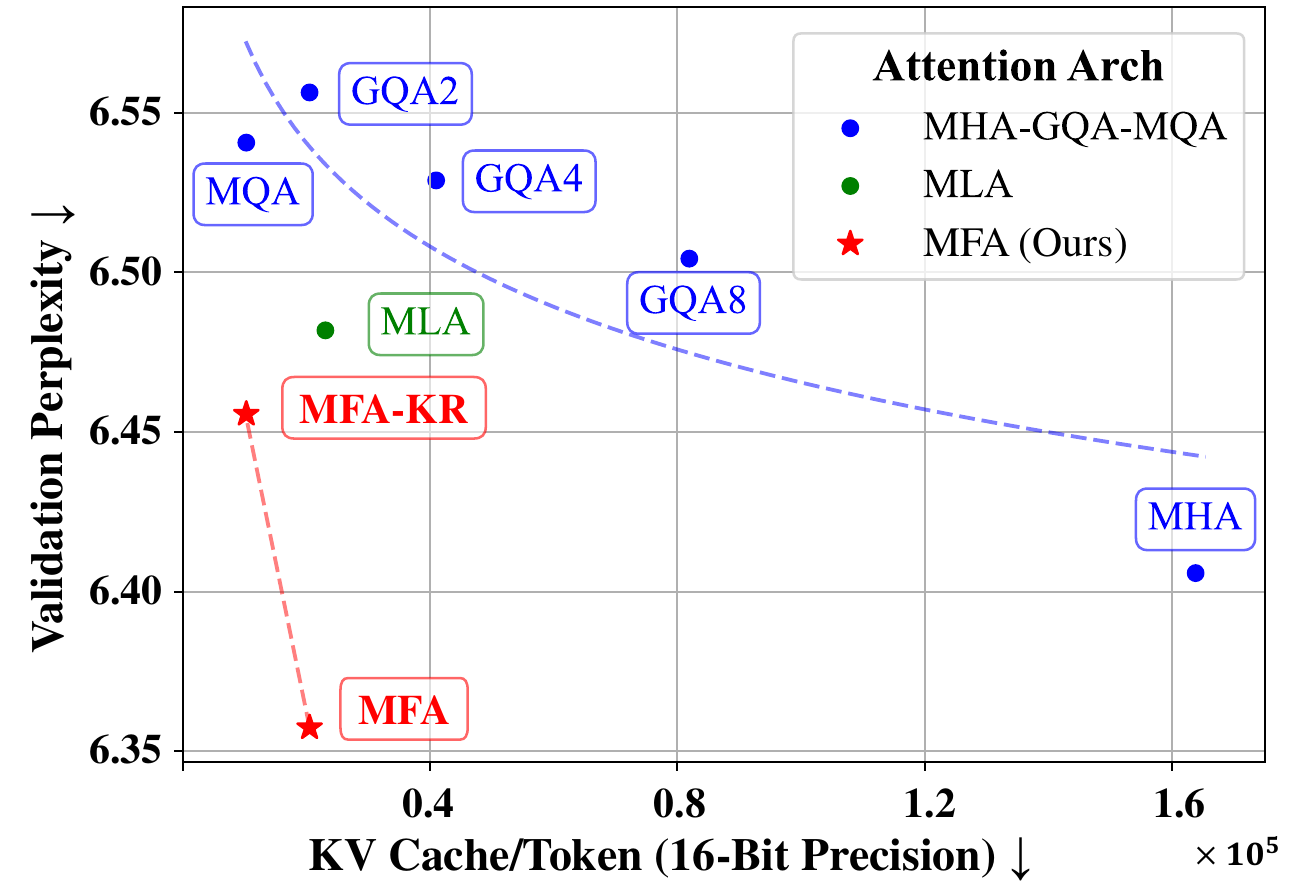}
    \vspace{-4ex}
    \caption{Validation perplexity vs. KV cache memory usage across different attention architectures in a 1B setting. KV Cache/Token indicates the KV cache size in bytes per token, assuming 16-bit precision for each element. Lower is better for both axes. }
    \label{fig:1b-sys-level}
    \vspace{-3ex}
\end{figure}

To address these challenges, Multi-Query Attention (MQA) and Grouped Query Attention (GQA) reduce KV cache usage by sharing key and value projections across heads \cite{shazeerFastTransformerDecoding2019, ainslieGQATrainingGeneralized2023}. 
Similarly, Multi-head Latent Attention (MLA) applies low-rank compression to key and value projections and only caches the latents \cite{deepseekai2024deepseekv2strongeconomicalefficient}. 
However, all methods fail to match MHA's performance under stringent KV cache budgets \cite{touvron2023llama2openfoundation}, as the added constraints on key and value projections limit the capacity of the attention module.

\begin{figure*}[ht]
    \centering
    \includegraphics[width=\linewidth]{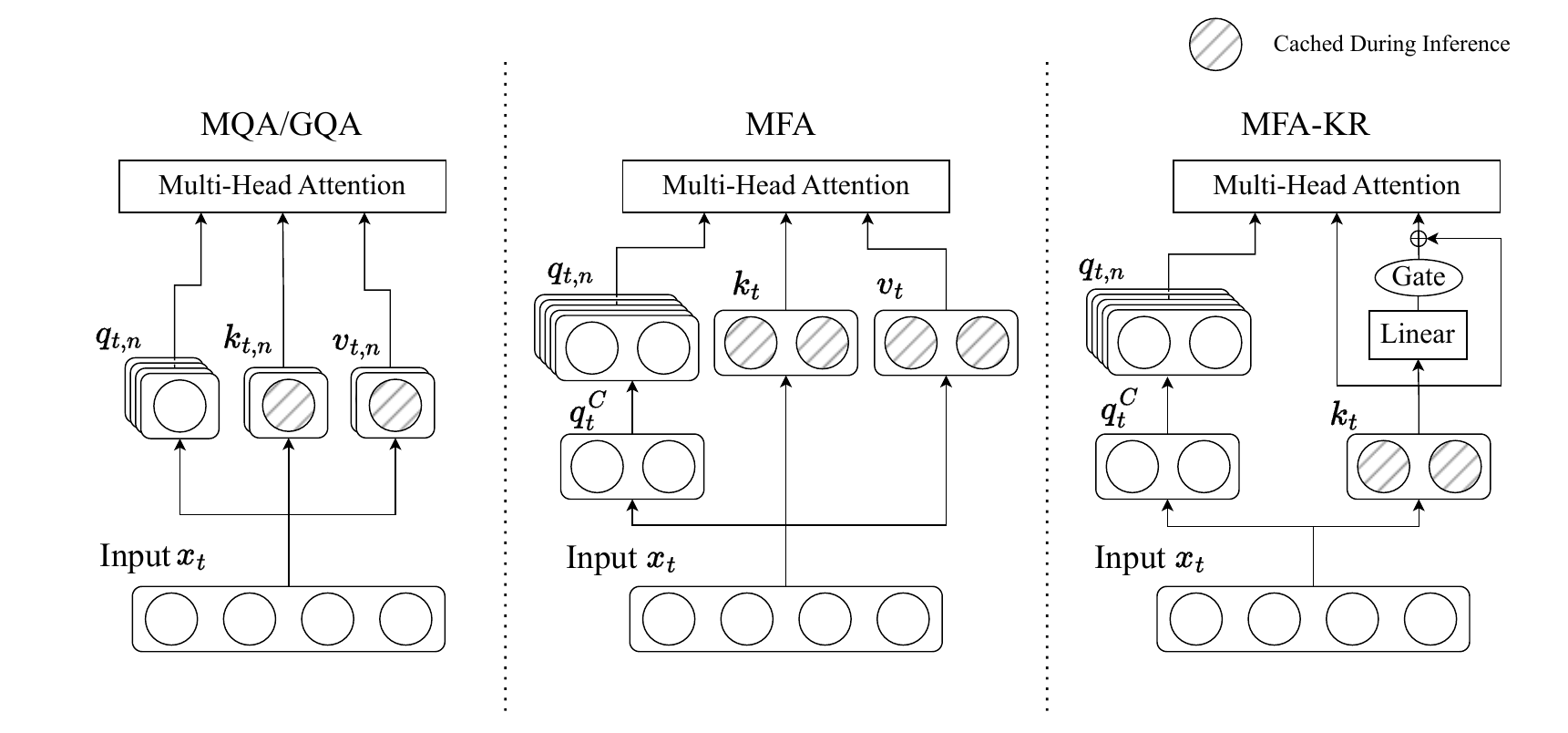}
    \vspace{-4ex}
    \caption{Simplified illustration of MFA/MFA-KR architecture compared with MQA/GQA architecture. By the expanding both the number and dimension of heads in a single-key-and-value manner, MFA/MFA-KR significantly enhances the model capacity under strict KV cache budget while maintaining parameter efficiency during scaling.}
    \label{fig:method}
\end{figure*}

Driven by these limitations, we analyze the modeling capacity in attention mechanisms and present a unified perspective on existing MHA variants. 
Our analysis reveals that the number and dimension of attention heads are critical for maintaining modeling capacity—an under-explored design aspect in current methods \cite{dubey2024llama, muennighoffOLMoEOpenMixtureExperts2024, jiangMixtralExperts2024a}.
This insight highlights the need to scale these factors efficiently to mitigate the capacity degradation caused by existing KV cache-saving techniques, pushing attention modules closer to their theoretical upper bound.

Inspired by this understanding, 
we propose \textbf{Multi-matrix Factorization Attention (MFA)}, a novel attention module, along with its variant, \textbf{MFA-Key-Reuse (MFA-KR)}. These attention modules are specifically designed to enhance modeling capacity under strict KV cache constraints.

Specifically, MFA employs a low-rank matrix factorization in the Query-Key (QK) circuit \cite{elhage2021mathematical}, enabling parameter-efficient scaling of both the number and dimension of heads without excessive kv cache usage. 
Building on MFA, MFA-KR reuses the key cache as value through a re-parameterized value projection with original key projection and a light weight gated projection. This minor modification cuts KV cache usage by an additional 50\% with negligible performance trade-offs.
Moreover, methods like MLA add complexity to support widely-adopted position embedding (i.e. RoPE), 
while our proposed MFA family naturally fit in current LLM training and inference ecosystems,
ensuring practical adoption without introducing additional architectural complexity.

We conduct extensive experiments to evaluate the performance and KV cache efficiency of MFA and MFA-KR, alongside detailed ablation studies on their design. 
Impressively, our proposed attention architecture is the only approach that performs comparably to standard MHA in terms of accuracy while adhering to strict KV cache constraints.
Specifically, in a 7B parameter model trained on 1T tokens, MFA and MFA-KR reduce KV cache usage by up to 93.7\% while achieving superior or comparable benchmark accuracies compared to MHA.

\section{Background: Capacity Analysis of Attention}
In order to delimit the scope of our analysis, we introduce the concept of Generalized Multi-Head Attention (GMHA).
It encompasses all multi-head mechanisms with linear query-key (QK) and value-output (VO) circuits, and per-head softmax attention. 
The QK circuit determines how information propagates between entities, and the VO circuit dictates how information is transformed \cite{elhage2021mathematical}. 
Fundamentally, GMHA can be described and analyzed using inference formulation and factorization formulation.
The inference formulation highlights how keys and values are computed and cached during inference, and factorization formulation clarifies the model’s capacity by interpreting QK and VO matrices as low-rank factorizations.
This offers a unified perspective on how different factorization strategies mediate the trade-off between model capacity and efficiency.

Within this framework, we identify Fully Parameterized Bilinear Attention (FPBA) as the upper bound of capacity.  
MHA and its variants can be regarded as a low-rank decomposition of FPBA, making FPBA a unified theoretical reference point for analysis.
Building on this understanding, we propose general design principles for constructing efficient and effective attention modules. 
These principles inform the design of the Multi-Matrix Factorization Attention (MFA) mechanism, which is introduced in the next section.

\subsection{Fully Parameterized Bilinear Attention}
Inspired by the work of \cite{shazeerTalkingHeadsAttention2020}, FPBA is defined as follows:
\begin{align}
    O_i = \sum_{c=1}^{H} \Big( \sum_{j=1}^{i} \phi\big( \frac{x_i W_{c} x_j}{\sqrt{H}} \big) x_j U_{c} \Big), 
    \label{GBA}
\end{align}
where $\phi$ denotes the softmax operator, $H$ is the embedding dimension, and $W_{c}$, $U_{c} \in \mathbb{R}^{H \times H}$ are independently parameterized for each channel $c$.
FPBA adheres to three key design principles to reach the theoretical maximum capacity within the GMHA framework. 
\romannumeral 1. Channel-specific interactions. In FPBA, each channel $c$ has a dedicated parameter $W_{c}$, the QK circuit $x_i W_{c} x_j$ captures channel-specific relations between $\mathbf{x}_i$ and $\mathbf{x}_j$.  
\romannumeral 2. The additivity of the \( c \)-th channels of \( x_i \) and \( x_j \) is generally not holding true. 
The VO circuit $x_j U_{c}$, which is fully parameterized as \( U \in \mathbb{R}^{H \times H \times H} \), and enables the projection of the \( H \)-dimensional embedding of \( x_j \) into arbitrary permutation of the \( H \)-dimensional embedding of \( x_i \);
\romannumeral 3. Full utilization of representations. FPBA fully utilizes the \(H\)-dimensional representations of both \(\mathbf{x}_i\) and \(\mathbf{x}_j\), without compressing any dimensions.
This flexibility allows unrestricted interactions across all dimensions, setting FPBA as the upper bound of capacity within the GMHA framework.

\subsection{Analysis of MHA and Its Variants}
As the prototypical instance of GMHA, MHA can be expressed using inference formulations~\eqref{MHAinference} and factorization formulations~\eqref{MHA}, as shown below.
\begin{align}
    O_i
    &= \sum_{c=1}^{n} \Big( \sum_{j=1}^{i} \phi(\frac{x_i Q_c (x_j K_{c})^T}{\sqrt{d}}) x_j V_{c} \Big) O_{c}^T \label{MHAinference} \\
    &= \sum_{c=1}^{n}\Big( \sum_{j=1}^{i} \phi(\frac{x_i ( Q_{c} K_{c}^T ) x_j^T}{\sqrt{d}}) x_j V_{c} O_{c}^T \Big) ,
    \label{MHA}
\end{align}
where $Q_c, K_c, V_c, O_c \in \mathbb{R}^{H \times d}$ represent the query, key, value, and output projections for $c$-th head. 
Comparing Eqs.~\eqref{GBA} and Eqs.~\eqref{MHA}, 
we can see that MHA is mathematically equivalent to a version of FPBA where $W_c$ and $U_c$ are approximated with low-rank factorization $Q_c K_c^T$ and $V_{c} O_{c}^T$ separately, both with bottleneck of $d$.
During inference time, by sharing parameters among $d$ channels rather than having a distinct set of parameters for each channel, and given that typically $n h = H$, the KV cache per token is reduced to $2H$. 

MQA extends the parameter sharing principles of MHA by using a single set of key and value parameters across all attention heads.
The formulations for MQA are nearly identical to those of MHA, with the key difference being that
$W_c$ and $U_c$ are factorized into $Q_c K^T$ and $V O_c^T$, 
where $K, V \in \mathbb{R}^{H \times d}$ are shared among all heads, each retaining rank $d$ but with shared parameter constraints. 
In inference formulations, by eliminating head-specific $K_c$ and $V_c$ parameters, 
the KV cache size is decreased to $2d$.

MLA adopts a more complex factorization of FPBA as follows:

\begin{small}
\begin{align}
    O_i
    &= \sum_{c=1}^{m} \big( \sum_{j=1}^{i} \phi(\frac{x_i S_q Q_c (x_j S_k K_{c})^T}{\sqrt{d}}) x_j S_v V_{c} \big) O_{c}^T \label{MLAinference} \\
    &= \sum_{c=1}^{m}\big( \sum_{j=1}^{i} \phi(\frac{x_i (S_q Q_c K_c^T S_k^T) x_j^T}{\sqrt{d}}) x_j S_v V_{c} O_{c}^T \big) ,
    \label{MLA}
\end{align}
\end{small}
where \( S_q, S_k, S_v \in \mathbb{R}^{H \times C} \) are shared among all heads, \( Q_c, K_c, V_c \in \mathbb{R}^{C \times d} \) are head-specific parameters, and \( C \) denotes the dimensionality of the latent factorization. We omit the decoupled RoPE design here for simplicity. 
Comparing factorization formulations Eq.~\eqref{MLA} with Eq.~\eqref{GBA}, 
it becomes evident that MLA employs parameter sharing across every \( H/m \) channels. 
Specifically, \( W_c \) is decomposed $S_q Q_c K_c^T S_k^T$, and 
\( U_c \) is decomposed $S_v V_{c} O_{c}^T$ . 
Although the intermediate dimension \( C > d\) typically, the overall rank remains to be the smallest dimension as $d$, without promoting the expressive capacity of the model.

\section{Multi-matrix Factorization Attention}
Building upon the analysis in the previous section, we arrive at the general design objective for efficient and effective attention module:
to find a matrix factorization scheme that minimizes parameter and KV cache size while pushing the model's capacity as close as possible to that of FPBA. 

\begin{table*}[htbp]
    
    \centering
\resizebox{0.99\linewidth}{!}{
    \begin{tabular}{l c c c c c c}
    
    \toprule
     {Method} &  {KV Cache} &  {Parameter} &  {Heads} &
    \makecell{{Factor. rank}\\ {per head}} & \makecell{{Shared latent}\\ {subspace Dim.}} & \makecell{Total \\ effec. rank}\\
    \midrule
    FPBA & $2H^2$ & $2H^3$ & $H$ & $H$ & $H$ & $H^2$\\
    MHA & $2H$ & $4H^2$ & $n$ & $d$ & $H$ & $nd$\\
    MQA & $2d$ & $(2+2/n)H^2$ & $n$ & $d$ & $d$ & $nd$  \\
    GQA & $2gd$ & $(2+2g/n)H^2$ & $n$ & $d$ & $gd$ & $nd$ \\
    MLA & $2C + d_r$ & \makecell{$H(3C+d_r+md)$\\$+mC(3d+d_r)$} & $m$ & $d$ & $C$ & $md$\\
    \midrule
    MFA & $2C$ & $H(3C + mC) + mC^2$ & $m$ & $C$ & $C$ & $mC$ \\
    \bottomrule
    \end{tabular}
    }
    \caption{
        Comparison of KV cache usage, parameter count, and total capacity, highlighting their capacity–efficiency trade-offs.
        \emph{Factor. rank per head} reflects each head’s factorization rank; 
        \emph{Shared latent subspace Dim.} indicates a common projection dimension across head’s factorizations; 
        \emph{Total effective rank} summing or combining ranks across all heads as total capacity approximates.
        Generally, $H > C > d = H/n$ and $m>n$. MFA achieves a higher \emph{Total Effective Rank} compared to other variants, making it the closest capacity approximation to FPBA.
    }
    \label{compare}
\end{table*}

Following these principles, we introduce Multi-matrix Factorization Attention (MFA), incorporating three key design strategies:
(1) Increasing the number and dimension of heads to minimize the amount of channel sharing in the propagation process and to provide greater expressive freedom for each head;
(2) Applying aggressive low-rank matrix factorizations on $W^n$ to enhance parameter efficiency as the model scales; 
(3) Utilizing single-key-and-value-head techniques to maintain minimal KV cache usage.

The inference and factorization expressions of MFA are given by:
\begin{align}
    O_i
    &= \sum_{c=1}^{n} ( \sum_{j=1}^{i} \phi(\frac{x_i S_q Q_c (x_j S_k)^T}{\sqrt{d}}) x_j S_v ) O_{c}^T \label{MFAinference} \\
    &= \sum_{c=1}^{n}( \sum_{j=1}^{i} \phi(\frac{x_i (S_q Q_c S_k^T) x_j^T}{\sqrt{d}}) x_j S_v O_{c}^T ) ,
    \label{MFA}
\end{align}
where \( S_q, S_k, S_v \in \mathbb{R}^{H \times C} \) are shared across heads, \( Q_c, O_c \in \mathbb{R}^{C \times C} \) are head-specific projection, and \( C \) denotes the low-rank factorization dimension.

During inference, as shown in Eq.~\eqref{MFAinference}, the key and value for each token \( x_j \) are calculated as \( x_j S_k \) and \( x_j S_v \) respectively, reducing the KV cache per token to \( 2C \). 
Compared to FPBA, the weight matrix \( W_c \) is decomposed into \( S_q Q_c S_k^T \), and the transformation matrix \( U_c \) is decomposed into \( S_v O_c^T \), both maintaining a rank of \( C \). 
This decomposition offers several advantages: (1)
\textbf{Scalable Head Count}: MFA allows for an increase in the number of heads with minimal parameter overhead ( $ \approx CH $ additional parameters per extra head). 
Moreover, the KV cache size remains constant regardless of the number of heads;  
(2) \textbf{Enhanced Head Expressiveness}: 
each head in MFA has a rank of \( C>d \) of others typically. 
This higher rank improves the expressive capacity of each head, allowing for more nuanced propagation and transmission; 
(3) \textbf{Compatibility with Positional Encodings}: unlike MLA, MFA seamlessly integrates with mainstream positional encodings such as Rotary Positional Encoding (RoPE), ensuring broader applicability across various transformer architectures.

To further optimize KV cache usage under stringent memory constraints, we introduce an extension of MFA called MFA-Key-Reuse (MFA-KR). This variant reuses the key cache by re-parameterizing the value projection based on the key projection, effectively reducing the KV cache size by an additional 50\%.
The re-parameterization is defined as:
\begin{align}
    S_v 
    &= S_k + \alpha \odot N S_k \\
    &= \left( I + \text{diag}(\alpha) N \right) W_K,  
\end{align}
where \( N \in \mathbb{R}^{C \times C} \), \( \alpha \in \mathbb{R}^{C} \), and \( \odot \) denotes element-wise multiplication. During training, the parameter \( \alpha \) is initialized as a zero vector to ensure that \( S_v \) equals \( S_k \) when training begins, because we empirically found it crucial for maintaining training stability.

To clarify the differences of MFA to other architectures, we present a straight-forward comparison in Table~\ref{compare}.
To maintain clarity, we omit MFA-KR and jointly key-value compressed version of MLA.
A GMHA model's capacity is influenced by two primary factors: Total Effective Rank (TER) and Shared Latent Subspace Dimension (SLSD). 
TER is defined as the product of the number of heads and the factorization rank per head (FRH), with higher TER indicating greater overall capacity. 
On the other hand, SLSD represents the dimension of the latent space shared across all heads.
A smaller SLSD reduces the KV cache size but constrains the model's capacity.
It is essential to note that the FRH must not exceed the SLSD, establishing a critical trade-off between capacity and efficiency.

\begin{table*}[t!]
\centering
\begin{tabular}{lccccc}
\toprule
 & \textbf{MHA}  & \textbf{MFA-KR} & \textbf{MFA} \\
\midrule
\# Activated Params  & 1.2B & 1.2B & 1.2B \\
\# Total Params & 6.9B & 6.9B & 6.9B \\
\textbf{KV Cache/Token} $\downarrow$ & 196.6K & \textbf{12.3K} & 24.6K   \\
\midrule
BBH \cite{suzgun2022challenging}     & 35.9                & 34.4          & \textbf{37.8}\\
MMLU \cite{hendrycks2020measuring}   & 45.2               & 43.5          & \textbf{45.5}\\
Hellaswag \cite{zellers2019hellaswag}  & \textbf{68.6}  & 67.5          & \textbf{68.6}\\
WG  \cite{sakaguchi2021winogrande}        & 60.2                 & \textbf{62.0} & 60.7         \\
BoolQ  \cite{clark2019boolq}     & 66.0              & 63.4          & \textbf{66.2}\\
PIQA \cite{bisk2020piqa}        & 76.0               & 75.5          & \textbf{77.0}\\
SIQA   \cite{sap2019socialiqa}     & 45.7               & 45.2          & \textbf{47.9}\\
SciQ   \cite{welbl2017crowdsourcingmultiplechoicescience}     & 71.6               & 68.8          & \textbf{74.3}\\
OBQA   \cite{OpenBookQA2018}    & 37.2               & 36.0          & \textbf{38.8}\\
Ruler  \cite{hsieh2024rulerwhatsrealcontext}     & 60.9              & 60.9           & \textbf{61.7} \\
DS1000 \cite{lai2022ds1000naturalreliablebenchmark}    & 11.2             & 11.0          & \textbf{12.1}\\
Math   \cite{hendrycks2021measuringmathematicalproblemsolving}      & 9.1                & 8.1           & \textbf{9.4} \\
\midrule
\textbf{Average Acc.} $\uparrow$ & 49.0 &  48.0 & \textbf{49.9} \\
\bottomrule
\end{tabular}
\caption{Benchmark accuracy and KV cache usage comparison among MFA, MFA-KR and MHA baseline. We scale the 7B model to 1 trillion training tokens, and MFA generally outperform MHA while using only 12.5\% of KV cache per token. MFA-KR demonstrates even less KV cache usage while compromising performance minimally. }
\label{tab:sys-level}
\end{table*}

As shown in Table~\ref{compare}, MFA achieves a higher TER compared to other methods, positioning it as the closest approximation to the theoretical upper-bound capacity represented by FPBA. Specifically, 
\romannumeral 1. Comparison with MQA: MFA achieves both a higher SLSD and a higher TER;  
\romannumeral 2. Comparison with MLA: under similar parameter budgets, MFA achieves a smaller KV cache size, a higher TER, and an equivalent SLSD;  
\romannumeral 3. Comparison with MHA: while MFA has a smaller SLSD than MHA, its TER is higher, leading to empirically superior results as shown in next section.

\section{Experiments}
\label{sec:exp}
We evaluate MFA for large language models from the following perspectives. 
First, we compare MFA and MFA-KR to MHA at 7B-scale MoE models with 1T training tokens on benchmark accuracies and KV cache usage. 
Second, we present the loss and KV cache curves of MFA and MFA-KR on increasing training scales. 
Third, we conduct extensive comparison with existing architecture variants and demonstrate the advantage of MFA and MFA-KR. 
Finally, we present studies on various design choices and validate the compatibility with different position embeddings.

\begin{figure*}[t!]
    \centering
    \includegraphics[width=\linewidth]{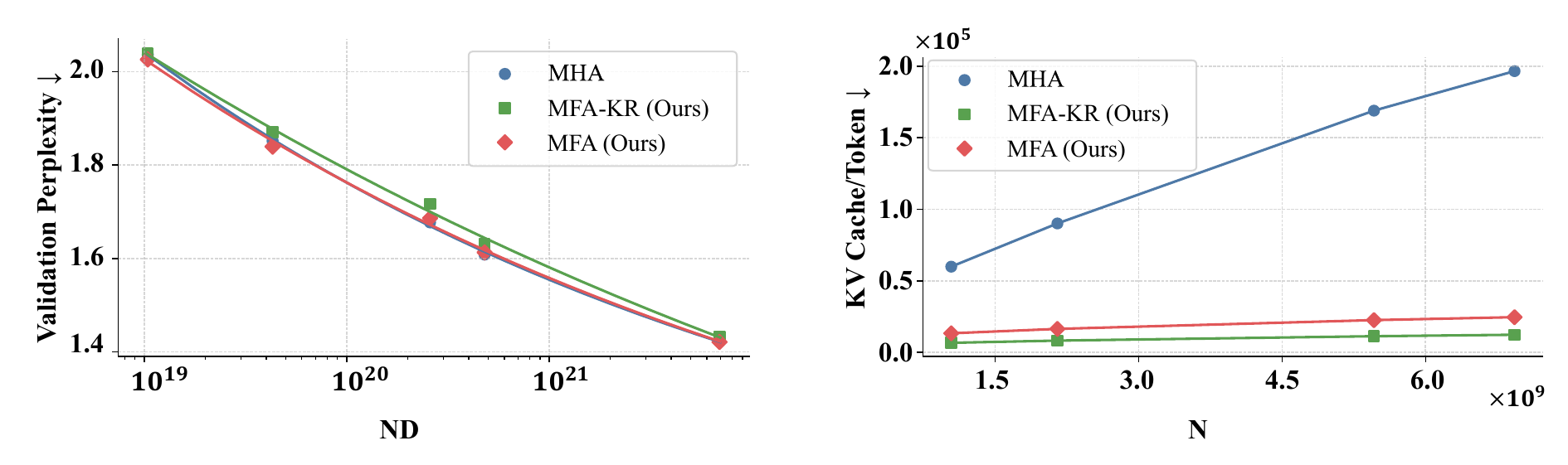}
    \caption{
    Scaling experiments among MHA, MFA, and MFA-KR. 
    \textbf{Left:} Loss vs. $ND$ scale, where $N$ denotes the total number of parameters and $D$ the total training tokens. MFA achieves comparable loss scaling curves to MHA. MFA-KR follows a similar scaling trend, albeit with a slight performance gap. 
    \textbf{Right:} KV cache usage per token vs. model size. 
    MFA and MFA-KR significantly reduce KV cache usage compared to MHA, with savings growing as model size increases.
    }
    \label{fig:mfa-scaling}
\end{figure*}
\subsection{Common Experimental Settings}
\label{sec:common-exp}
In all our experiments, we train our models with language modeling loss on a high-quality training data corpus created internally, including web text, mathematical material, and code, tokenized using the BPE \cite{sennrich2015neural} tokenizer with vocabulary size of 65536. 
We adopt pre-normalization using RMSNorm \cite{zhang2019root}, SwiGLU \cite{shazeer2020glu} activation function for FFN without dropout, and rotary position embeddings \cite{su2024roformer} with base frequency set to 500,000 \cite{dubey2024llama}. 
All models are trained from scratch and the weights are initialized in the following method: all weights of linear layers are first initialized from a truncated normal distribution with mean zero and standard deviation 0.02, and then for the output projection of attention and the $W_2$ of the GLU we divide the initialized value by $\sqrt{2 \cdot \text{layer\_idx}}$, which is adopted from \cite{narayananEfficientLargeScaleLanguage2021}.

All models are trained with AdamW \cite{loshchilovDecoupledWeightDecay2019} optimizer, with $\beta$ = [0.9, 0.95], eps=$10^{-8}$, weight decay factor of 0.1 and gradient clipping norm of 1.0. 
For learning rate schedules, we use a linear warmup for the first $2000$ steps and a cosine decay to $10^{-5}$ for the remainder of training. We set the sequence length to 16384 tokens. We hold out a validation set of $\approx$ 10M tokens drawn from the same distribution of training data for evaluation purposes.

\subsection{Language Modeling Evaluation}
\label{sec:lm-eval}
We train MoE language models with 7B total parameters and 1B activated parameters on 1T tokens to compare MFA/MFA-KR with the MHA. 

\paragraph{Setup.}
We adopt a modified version of DeepSeekMoE \cite{dai-etal-2024-deepseekmoe} as basic architecture, including shared experts and the first layer using dense FFN, but using coarse-grained experts due to system efficiency considerations.
We align hidden size, layers, the total number and the activated number for parameters across all models.
FFN dimensions of the first dense layer, total number and activated number of experts are slightly adjusted to meet the requirements. 
We employ an expert-level load balance loss \cite{shazeer2017outrageously}, with load balance factor as 0.01.
All models are trained with same peak learning rate as $8.4\times10^{-4}$ and the each training batch contains 7.3 million tokens, and the training spans 140K steps, totaling 1 trillion tokens.
More details can be found in Appendix~\ref{app:hp}.

For evaluation, we benchmark the models on various downstream tasks within a unified evaluation framework. We select extensive tasks, including reasoning, knowledge, and factual accuracy, providing a holistic assessment of model performance.

\paragraph{Results.}
Table~\ref{tab:sys-level} presents a comparison of MFA, MFA-KR, and MHA on downstream language modeling tasks. 
The results show that MFA achieves superior average benchmark accuracy (49.9\%) compared to MHA (49.0\%), while reducing KV cache usage per token by 87.5\% (from 196.6KB to 24.6KB). This highlights MFA’s ability to balance strong modeling capacity while maintaining exceptionally low KV cache memory usage. 
MFA-KR further minimizes KV cache usage to just 12.3KB per token—only 6.25\% of MHA’s storage needs—by reusing key caches as values. While MFA-KR incurs a slight accuracy trade-off, it remains competitive and is well-suited for scenarios where memory constraints are paramount.

\begin{table*}[t!]
\centering
\begin{tabular}{ccccc}
\toprule
\textbf{Factor.} &  $d \cdot n$ & \textbf{\# Params} & \textbf{KV Cache/Token} \textdownarrow & \textbf{Val PPL} \textdownarrow \\
\midrule
\ding{55}   & $H$     & 1.08B & 20K & 6.54          \\
\ding{55}   & $1.75H$ & 1.20B & 20K & 6.38          \\
\checkmark  & $1.75H$ & 1.08B & 20K & \textbf{6.36} \\
\bottomrule
\end{tabular}
\caption{Effect of QK circuit factorization on scaling $d \cdot n$ in MFA. Without factorization, increasing $d \cdot n$ improves validation perplexity increases parameter count. 
Factorization allows MFA to match validation perplexity while maintaining parameter efficiency, enabling parameter-efficient scaling of $d$ and $n$. Factor. and C./T. represents factorization and the KV cache/token.}
\label{tab:qk-factorization}
\end{table*}

\subsection{Scalabiliy Experiments}
\label{sec:scaling}
We compare the loss scaling curves between MHA, MFA and MFA-KR. The scaling law is supposed to extrapolate the performance at larger scales. 
\paragraph{Setup.}
We use the same model architecture setup mentioned in Section~\ref{sec:lm-eval}. 
We train MoE language models of various sizes (i.e., 1.0B, 2.1B, 5.5B, 6.9B) and various numbers of tokens (i.e., 10B, 20B, 48B, 69B) while keeping the model sparsity (the ratio of activated number of parameters to total number of parameters) constant. 
We also add our 7B model with 1T training token experiments in our scaling curve results. 
We use loss on our valuation set as the evaluation metric. 
More details are shown in Appendix~\ref{app:scaling}.  
\paragraph{Results.}

We compare the scalability and efficiency of MFA and MFA-KR with MHA through loss scaling across various model sizes and training tokens, as shown in Figure~\ref{fig:mfa-scaling}. The left plot shows validation loss curves with respect to $ND$ scale (where $N$ is the number of parameters and $D$ the total training tokens). MFA matches MHA’s loss scaling behavior, confirming its strong modeling capacity, while MFA-KR demonstrates a similar trend with a minor performance gap, making it suitable for highly memory-constrained scenarios.

The right plot compares KV cache usage per token across model sizes. At largest scale, MFA reduces KV cache requirements by 87.5\% compared to MHA, with MFA-KR achieving even greater savings at just 6.25\% of MHA’s usage. The relative savings grow with larger model sizes, highlighting the scalability of both MFA and MFA-KR methods.

\subsection{Ablation Study}
\label{sec:abl}
We conduct ablation study on 1B-scale dense model. 
We set hidden size to 2048, number of layers to 20, and keep the total number of parameter the same by adjusting the FFN size for different attention architectures unless otherwise stated. 
All models are trained with peak learning rate as $9.63\times10^{-4}$, and each training batch contains 0.4M tokens. Total training steps are set to 50k. Therefore, all models consume 20B tokens in training. Ablation studies quantifies the accuracy of models using perplexity evaluated on validation set.

\paragraph{Comparing with Other Attention Architectures.}
We show the trade-offs between validation perplexity and KV cache usage across various attention architectures in our 1B dense model setting in Figure~\ref{fig:1b-sys-level}. 
The comparison includes MHA, GQA, MQA, and MLA, representing a spectrum of design choices for balancing modeling capacity and memory efficiency. 
Models in the MHA-GQA-MQA spectrum reflect the baseline trade-offs for achievable accuracy and memory usage. 
More recent MLA architecture is also evaluated in our setting. 
All models have undergone the same training recipe, except that MLA uses the initialization method mentioned in \cite{deepseekai2024deepseekv2strongeconomicalefficient}. 
We find that MLA is quite sensitive to the initialization method and only with this initialization can it achieve reasonable performance. Details are elaborated in Appendix~\ref{app:init-mla}.
Our implementation and architecture hyperparameter of MLA refers to the open-source model DeepSeek-V2-Lite. 

The results demonstrate that MFA and MFA-KR achieve a new Pareto frontier for accuracy and memory trade-offs. MFA achieves the lowest validation perplexity while using only 12.5\% of KV cache memory compared to MHA. MFA-KR further reduces KV cache usage while maintaining competitive accuracy.
Notably, MFA and MFA-KR outperform MLA and the MHA-GQA-MQA baselines in terms of both validation perplexity and KV cache efficiency, and MFA achieves even better performance compared to MHA baseline.

\paragraph{Efficiently Scale up $d$ and $n$.}
To evaluate the parameter efficiency of scaling $d \cdot n$ in MFA, we ablate over QK circuit factorization, as shown in Table~\ref{tab:qk-factorization}. 
First we show that without factorization design, increasing $d \cdot n$ from $H$ to $1.75H$ improves validation perplexity, enhancing the model capacity under strict KV cache usage in this setting. 
However, vanilla scaling up $d$ and $n$ comes at the cost of higher parameter count (10\% more in this setting). 
In contrast, applying factorization allows MFA to scale $d \cdot n$ to $1.75H$ while keeping the parameter count fixed. 
This approach achieves the as good validation perplexity, highlighting that the factorization in MFA enables the parameter-efficient scaling of $d$ and $n$.

\begin{table*}[t!]
\centering
\begin{tabular}{lcc}
\toprule
\textbf{Architecture} & \textbf{KV Cache/Token} \textdownarrow & \textbf{Val PPL} \textdownarrow  \\
\midrule
MHA                                & 163K  & 6.41    \\ 
MFA                                & 20K   & 6.35    \\
\hspace{0.5em} +vanilla KR         & 10K   & 6.55    \\
\hspace{0.5em} +extra value proj.  & 10K   & 7.88    \\
\hspace{0.5em} +residual connect   & 10K   & 6.65    \\
\hspace{0.5em} +gating = MFA-KR    & 10K   & 6.45    \\
\bottomrule
\end{tabular}
\caption{Ablation study for how to arrive at current MFA-KR architecture design choice. KV Cache/Token indicates the KV cache size in bytes per token, assuming 16-bit precision for each element. }
\label{tab:kv-share-ablation}
\end{table*}

\begin{table*}[t!]
\centering
\begin{tabular}{lcc}
\toprule
\textbf{Architecture} & \textbf{KV Cache/Token} $\downarrow$ &  \textbf{Val PPL} $\downarrow$   \\
\midrule
MHA      &  163K          & 6.60     \\
MFA-KR   &  \textbf{10K}  & 6.48     \\
MFA      &  20K           & \textbf{6.45}     \\
\bottomrule
\end{tabular}
\caption{Performance of MFA and MFA-KR compared to MHA with ALiBi as the positional embedding.}
\label{tab:alibi}
\end{table*}
\paragraph{Design Choices for Key-Reuse.}
We ablate the design choice for MFA-KR, as shown in Table~\ref{tab:kv-share-ablation}. 
Starting from MFA, we incrementally test key reuse and additional design improvements. Vanilla key reusing strategy incur non-negligible performance drop. While adding extra value projection aims to enhance modeling capacity, it suffers from training instability and gets bad performance. 
Adding a residual connection mitigates instability but still results in suboptimal performance. 
Finally, incorporating a zero-initialized gating mechanism addresses both stability and performance issues, resulting in MFA-KR, which matches MHA’s performance while further halving the KV cache usage compared to MFA.

\paragraph{Different Position Embeddings}
ALiBi \cite{press2021train} is also a common position embedding \cite{almazroueiFalconSeriesOpen2023} with built-in zero-shot length extrapolation ability. Table~\ref{tab:alibi} shows that MFA and MFA-KR maintain advantage with changed position embedding.

\section{Related Works}
Notable efforts have focused on architectural modifications to minimize KV cache usage besides MQA, GQA and MLA which we elaborated in previous section. 
CLA\cite{brandonReducingTransformerKeyValue2024} and MLKV\cite{zuhri2024mlkvmultilayerkeyvalueheads} attempt to share key and value between layers, further reducing KV cache memory storage overhead. However, since even shared KV cache must be re-loaded in each layers seperately, this method does not reduce the KV cache memory traffic, thus having no effect on the latency for core attention computation.

Other works aim to replace all or part of Softmax Attention operations with alternatives that maintain a constant cache state size relative to sequence length, such as SSMs \cite{gu2024mambalineartimesequencemodeling, lieber2024jambahybridtransformermambalanguage} or linear attention\cite{katharopoulos2020transformers,pengEagleFinchRWKV2024}. This reduces the cache state size significantly in extremely long-context region, and can be combined with our proposed MFA and MFA-KR in hybrid manner.

Another active area of research seeks to boost the capacity of attention modules. \cite{bhojanapalli2020lowrank} indentifies the dimension of each head in MHA bottlenecks the capacity of attention module, and the situation may become worse if adhere to current high weight decay training recipe \cite{kobayashi2024weight}. Other works like Talking-Head Attention \cite{shazeerTalkingHeadsAttention2020} and DCMHA \cite{xiaoImprovingTransformersDynamically2024} try to enable information exchange between heads to augment model capacity. Though potential performance gain can be achieved, this modification is not compatible with commonly used Flash Attention \cite{dao2022flashattention}, limiting the scaling up of these architectures. 

Parameter efficiency in transformer models has been extensively studied, especially in finetuning domain \cite{hu2021lora}. There are also works focusing on pretraining parameter efficiency like LPA\cite{lvScalableEfficientTraining2024}; however, they do not investigate the effects under limited KV cache budget.

\section{Conclusions}
We present Multi-matrix Factorization Attention (MFA) and its variant MFA-Key-Reuse (MFA-KR) as scalable solutions to achieve superior performance while drastically reducing KV cache requirements. Our experiments demonstrate that MFA achieves superior benchmark accuracies with up to 87.5\% less KV cache compared to MHA, while MFA-KR pushes memory efficiency further by halving KV cache requirements with minimal trade-offs. 

\newpage

\section*{Limitations}

We do not directly evaluate the system-level implications of KV cache reduction, such as its impact on end-to-end inference efficiency for large-scale, long-context models. 
The integration of MFA with other architectural innovations, such as CLA or linear attention mechanisms, is not explored. Investigating these combinations could further optimize memory usage and performance, particularly for resource-constrained environments with high model capacity requirements. 
Moreover, we have not validated the performance of MFA and MFA-KR at even larger scale.

\section*{Acknowledgments}
The work was supported by National Science and Technology Major Project of China (2023ZD0121300).

\bibliography{anthology,custom}
\appendix
\section{Details of Hyper-Parameters}
\label{app:hp}
In this section, we provide more elaboration on the implementation details for Section~\ref{sec:exp}.

\begin{table*}[t!]
\centering
\resizebox{0.95\linewidth}{!}{
\begin{tabular}{lcccc}
\toprule
\textbf{Design Choice} & \textbf{Extra Op. for RoPE} & \textbf{\# Params (B)} & \textbf{Cache/Token (K/V)} $\downarrow$ & \textbf{Perplexity} $\downarrow$ \\
\midrule
\textbf{Key Projection} & & & & \\ 
Compressed key         & \checkmark    & 1.12 & 12.8K \textit{(K)} & 6.36 \\
Single key     & \ding{55}     & 1.06 & 5.1K \textit{(K)} & \textbf{6.34} \\
\midrule
\textbf{Value Projection} & & & & \\
Compressed value          & \ding{55}     & 1.07 & 10.2K \textit{(V)} & 6.60 \\
Single value     & \ding{55}     & 1.06 & 5.1K \textit{(V)} & \textbf{6.33} \\
\bottomrule
\end{tabular}
}
\caption{Comparison of key and value projection designs used in attention modules. The column \textit{Cache/Token (K/V)} indicates the size of the key cache (\textit{K}) or value cache (\textit{V}) per token in bytes, assuming 16-bit precision. \textit{Compressed key/value} represent the designs adopted in MLA, where "compressed key" requires extra operations for RoPE. \textit{Single key/value} represent the design adopted in MFA. In key/value projection experiments, the value/key projections use the same multi-head implementations to keep fair comparisons. }
\label{tab:kv_projection_combined}
\end{table*}

\subsection{Common experimental Settings}
The training data we use in our experiments has gone through thorough cleaning procedure, minimizing the harmful contents and personal information about private individuals. 
Data is sampled using Best-Fit-Packing \cite{ding2024fewer} with bin size of 8 to mitigate truncation issues without disturbing the data distribution. Samplings are conducted with fixed random seed to ensure fairness when we compare different model architectures. We conduct preliminary experiments to test the validation perplexity fluctuation under same training and evaluation procedure, and find that the standard deviation of validation perplexity is smaller than 0.005. Therefore all our experiments only conduct the training once and apply the standard evaluation protocols. We perform all experiments using PyTorch \cite{ansel2024pytorch}, and the usage adheres to the PyTorch License. 

\subsection{Language Modeling Evaluation}
\label{app:lm-eval}
We present the model hyperparameter for 7B MoE models used in language model evaluation experiment in Table~\ref{tab:app-sys-level-setting}. 

\begin{table}[ht]
\centering
\resizebox{0.99\linewidth}{!}{
\begin{tabular}{lccc}
\toprule
\textbf{Architecture} & \textbf{MHA} & \textbf{MFA} & \textbf{MFA-KR} \\
\midrule
\# params (B)             & 6.9 & 6.9 & 6.9 \\
\# act. params (B)   & 1.2 & 1.2 & 1.2 \\
\midrule
Hidden Size     & 2048       & 2048 & 2048 \\
Layers          & 24         & 24   & 24 \\
$n$           & 16         & 18   & 18 \\
$d$           & 128        & 256  & 256 \\
\# Experts      & 33         & 29   & 29 \\
MoE Top-k           & 2          & 2    & 2 \\
MoE FFN Size    & 1312       & 1504 & 1536 \\
Share FFN Size  & 2624       & 3008 & 3016 \\
\bottomrule
\end{tabular}
}
\caption{Architectural hyperparameters for language model evaluation experiment. }
\label{tab:app-sys-level-setting}
\end{table}

\subsection{Scalability Experiments}
\label{app:scaling}
We present model and training hyperparameter details for scalability experiments. 

\begin{table}[ht]
\centering
\resizebox{0.99\linewidth}{!}{
\begin{tabular}{lcccc}
\toprule
\textbf{Exp. Settings} & 1B & 2B & 5B & 7B \\
\midrule
\textbf{\# params (B)}           & 1.0      & 2.2      & 5.5      & 6.9      \\
\textbf{\# act. params (B)} & 0.2      & 0.4      & 0.9      & 1.2      \\
\textbf{Train Tokens(B)}         & 10       & 20       & 47       & 69       \\
\midrule
\textbf{Hidden Size}             & 1152     & 1408     & 1920     & 2048     \\
\textbf{Layers}                  & 13       & 16       & 22       & 24       \\
\textbf{$n$}          & 9        & 11       & 15       & 16       \\
\textbf{$d$}                & 128      & 128      & 128      & 128      \\
\midrule
\textbf{Learning Rate}           & 8.0e-4 & 5.9e-4 & 4.0e-04 & 3.7e-4 \\
\textbf{Batch Size (M)}          & 0.3      & 0.4     & 1.6    & 0.8       \\
\bottomrule
\end{tabular}
}
\caption{Common hyperparameters for scaibility experiments at each scaling setting. }
\end{table}

\subsection{Ablation Study}
\label{app:abl}
We present the detailed model architecture hyperparameters used in our ablation study, including Feed-Forward Network (FFN) size, the number of attention heads \( n \), and the head dimension \( d \), as shown in Table~\ref{tab:abl_hyperparams}. The low rank dimension for MLA is set to 512, and the dimention with RoPE are set to 64, following DeepSeek-V2-Lite.

\begin{table}[ht]
    \centering
    \begin{tabular}{lccc}
        \toprule
        \textbf{Architecture} & \textbf{FFN Size} & \textbf{\( n \)} & \textbf{\( d \)} \\
        \midrule
        MHA        & 6008  & 16 & 128 \\
        GQA8       & 6680  & 16 & 128 \\
        GQA4       & 7032  & 16 & 128 \\
        GQA2       & 7200  & 16 & 128 \\
        MQA        & 7304  & 16 & 128 \\
        MLA        & 6504  & 16 & 128 \\
        \midrule
        MFA        & 7168  & 14 & 256 \\
        MFA-KR     & 7232  & 14 & 256 \\
        \bottomrule
    \end{tabular}
    
    \caption{Model architecture hyperparameters for the ablation study. The table includes Feed-Forward Network (FFN) sizes, the number of attention heads (\( n \)), and head dimensions (\( d \)) for different attention architectures.}
    \label{tab:abl_hyperparams}
\end{table}

\subsection{Key and Value Projection Design Ablation}
We ablate the single-key-and-value design choice used in MFA, comparing it to the compressed-key-and-value design adopted in MLA. Results are shown in Table~\ref{tab:kv_projection_combined}. 
For key projections, the compressed design requires additional operations to ensure compatibility with RoPE. In contrast, the single-key design used in MFA eliminates the need for additional RoPE related modifications, achieving simplicity and robustness. 
For value projections, compressed-value designs incur a notable performance degradation compared to single-value designs adopted in MFA. 
These findings validate the design choices in MFA, where both single-key-and-value head design choices align with the core design principles of MFA, which prioritize practicality and modeling capacity under strict KV cache constraints.

\section{Initialization Study on MLA}
\label{app:init-mla}
In our experiments, we find that MLA is sensitive to the initialization method, performing poorly under our default setting, while MHA and MFA remain robust across different initializations. We leave the investigation for the underlying reasons as interesting future work. Results are summarized in Table~\ref{tab:mla-init}.

\newpage
\begin{table}[ht]
    \centering
    \resizebox{0.99\linewidth}{!}{
    \begin{tabular}{lcc}
        \toprule
        \textbf{Architecture} & \textbf{Initialization} & \textbf{Val PPL} $\downarrow$ \\
        \midrule
        \multirow{2}{*}{MLA} & Ours      & 6.73 \\
                             & DeepSeek  & 6.48 \\
        \midrule
        \multirow{2}{*}{MHA} & Ours      & 6.41 \\
                             & DeepSeek  & 6.44 \\
        \midrule
        \multirow{2}{*}{MFA} &  Ours      & 6.36 \\
                             & DeepSeek  & 6.43 \\
        \bottomrule
    \end{tabular}
    }
    \caption{Validation perplexity of MLA, MHA, and MFA under different initialization methods. MLA shows sensitivity to initialization, with a large performance gap between our and DeepSeek's method. In contrast, MHA and MFA exhibit robust performance across both initialization settings.}
    \label{tab:mla-init}
\end{table}

\section{Potential Risks}
\label{app:risks}
Although we conduct detailed processing to filter harmful content, the pretrain models we study can still generate harmful or biased content due to its unaligned nature.

\end{document}